\documentclass[11pt]{article}
\usepackage{acl2016}
\usepackage{times}
\usepackage{url}
\usepackage{latexsym}

\usepackage{graphicx}
\usepackage{bm}
\usepackage{multirow}
\usepackage{pifont}
\usepackage{amsmath} 
\usepackage{amsthm}
\usepackage{algorithm}
\usepackage{algpseudocode}
\usepackage{varwidth}
\newcommand{\cmark}{\ding{51}}
\newcommand{\xmark}{\ding{55}}
\newtheorem{theorem}{Theorem}

\algnewcommand{\LeftComment}[1]{\Statex \(\triangleright\) #1}

\aclfinalcopy 

\title{A FOFE-based Local Detection Approach for Named Entity Recognition and Mention Detection}

\author{Mingbin Xu \and Hui Jiang \\
  Department of Electrical Engineering and Computer Science \\
  Lassonde School of Engineering, York University \\
  4700 Keele Street, Toronto, Ontario, Canada\\
  {\tt xmb@cse.yorku.ca} \and  {\tt hj@cse.yorku.ca}  
 }

\date{}

\begin{document}
\maketitle

\begin{abstract}
In this paper, we study a novel approach for named entity recognition (NER) and mention detection in natural language processing. Instead of treating NER as a sequence labelling problem, we propose a new local detection approach, which rely on the recent fixed-size ordinally forgetting encoding (FOFE) method to fully encode each sentence fragment and its left/right contexts into a fixed-size representation. Afterwards, a simple feedforward neural network is used to reject or predict entity label for each individual fragment. The proposed method has been evaluated in several popular NER and mention detection tasks, including the CoNLL 2003 NER task and  TAC-KBP2015 and TAC-KBP2016 Tri-lingual Entity Discovery and Linking (EDL) tasks. Our methods have yielded pretty strong performance in all of these examined tasks. This local detection approach has shown many advantages over the traditional sequence labelling  methods.
\end{abstract}

\section{Introduction}

Natural language understanding is an important task in artificial intelligence. Natural language processing (NLP) has been extensively studied for many decades. The conventional NLP techniques include the rule-based symbolic approaches widely used about 20 years ago, and the more recent statistic approaches that rely on feature engineering and relatively simple statistical models, such as conditional random fields (CRFs). In the past few years, neural networks based deep learning approaches have achieved huge successes in many other applications, ranging from speech recognition to image classification. These approaches are drawing more and more attention in the NLP community. 

Among many different NLP problems, in this paper, we are interested in a fundamental problem in NLP, namely named entity recognition (NER) and mention detection. Named entity recognition (NER) and mention detection are a very  challenging task in NLP, laying the foundation of almost every NLP application.
NER and mention detection is a task of identifying entities (named and/or nominal) from raw text, and classifying  the detected entities into one of pre-defined categories such as person, organization, location, etc.  
It is a core component of almost every other higher level NLP tasks, such as information extraction, language understanding, knowledge base population.  For example, given a raw text of English like {\it ``S.E.C. chief Mary Shapiro left Washington in December.''}. The NER task is to detect and label all the mentioned entities with correct  categories, such as
\begin{quote}
	\label{emp:3types}
	\small
	${[S.E.C.]}_{ORG}$ chief ${[Mary\ Shapiro]}_{PER}$ left ${[Washington]}_{LOC}$ in December .
\end{quote}

In many applications, we may need to detect not only named entities but also the nominal mentions. For example, in an English sentence like {\it ``Mark and his closest friend Scarlet, a cello player, joined the same company.''}, we may want to label both named and nominal entities, which are important for other NLP tasks such as coreference resolution. 

\begin{quote}
	\label{emp:3types}
	\small
	$[Mark]_{PER}$ and his closest $[friend]_{PER\_N}$ $[Scarlet]_{PER}$, a cello $[player]_{PER\_N}$, joined the same music $[company]_{ORG\_N}$.
\end{quote}

In other applications, we may need to detect all entities and mentions, which are even nested or embedded. For example, in an English sentence like {``\it He used to study in University of Toronto.''}, where {\it Toronto} is a LOC entity, embedded in another longer ORG entity {\it University of Toronto}.

\begin{quote}
	\label{emp:nested-ex}
	\small
	He used to study in \\  ${[University\ of\ {[Toronto]}_{LOC}]}_{ORG}$.
\end{quote}

Traditionally, like many other NLP problems, NER and mention detection are normally formulated as a sequence labelling problem, where a tag is sequentially assigned to each word in the input sentence one by one. Depending on how these tags are defined, we may solve many NLP problems, such as chunking, part-of-speech (POS) tagging, NER, semantic parsing and so on. The sequence labelling has been extensively studied in the NLP community. The core problem in sequence labelling is to model the conditional probability of an output sequence given an arbitrary input sequence. Traditionally, many hand-crafted features are combined with statistical models, such as conditional random fields (CRFs), to compute conditional probabilities. More recently, some popular neural networks, including convolutional neural networks (CNNs), recurrent neural networks (RNNs) and LSTMs, are proposed to solve these sequence modeling problems under the popular sequence to sequence modelling framework. The relevant work will be briefly reviewed in Section \ref{sec_related_work}. 
In the test stage, given any input sequence, the learned models are used to compute the conditional probabilities and the output sequence is generated by the well-known Viterbi decoding algorithm. 

In this paper, we propose to use a novel local detection approach to solve NER and mention detection problems.  The idea can be easily extended to many other sequence labelling problems in NLP.  In our proposed methods, instead of globally modelling the whole sequence in training and jointly decode the entire output sequence in test, 
our methods will locally judge and verify every possible fragment in a sentence for the possible label based on the underlying fragment itself as well as its left and right contexts in the sentence.  Taking NER as example, our method will examine all word segments (up to a certain length) in a sentence one by one. At each time, a word segment will be examined individually based on the underlying segment itself and its left and right contexts in the sentence to determine whether this word segment is a valid named entity. If yes, the model will output the category for this entity as well. Otherwise, the segment is rejected and no NER tag is generated for this segment.
This approach more or less conforms to the way human is resolving an NER problem. Given any word fragment and its contexts in a sentence or paragraph, people normally can pretty accurately determine whether this word segment is a named entity or not. People rarely need to conduct a global decoding over the entire sentence to make such a decision. 
The key to making an accurate local decision for each individual fragment is to have a full access to the fragment itself as well as its complete contextual information. 
The main pitfall to implement this idea is that we can not easily encode the segment and its contexts in models since they are all variable-length sequences in natural languages. Many feature engineering techniques have been proposed for this but all of these methods will inevitably lead to information loss. 
In this work, we propose to use a recent fixed-size encoding method, namely fixed-size ordinally forgetting encoding (FOFE) \cite{zhang2015fixed}, to solve this problem.  The FOFE method is a simple recursively encoding method for any variable-length sequence. There is a nice theoretical property to guarantee that FOFE codes can almost uniquely encode
any variable-length sequence of words into a fixed-size representation without losing any information. Here, we propose to use the FOFE methods to fully encode the left and right contexts for each word segment, and then a simple feedforward neural network can be trained to make a precise recognition for each individual word segment based on the fixed-size presentation of the contextual information. This FOFE-based local detection approach is more appealing to NER and mention detection. First of all,  we may be able to totally get rid of feature engineering in these NLP problems since FOFE only relies on a single forgetting factor to fully encode any sequence. Second,  we can easily handle some difficult problems in NER, e.g. nested and embedded entities labels, under this local detection framework  without too much modification. Next,  this local detection approach can make better use of partially labelled data available from many application scenarios. For sequence labelling models, we need to label all entities in a sentence. It is always expensive to fully label all sentences. In some cases, if only some (not all) entities are labelled, it is not very effective to learn a sequence labelling model based on this type of data. However, it is quite different for the local detection approach. Every single labeled entity, along with its left and right contexts, may be used to learn the model. At last, due to the flexible encoding strategy by FOFE, we may rely on some simple neural networks for recognition, such as plain feedforward fully-connected neural networks. These models are much faster to train and easier to tune. In the test stage, all possible word segments from a sentence may be packed into a mini-batch, which can be jointly recognized in parallel on GPUs. This leads to a very fast decoding process as well.

In this paper, we have applied this FOFE-based local detection approach to several popular NER and mention detection tasks, including the CoNLL 2003 NER task and 
TAC-KBP2015 and TAC-KBP2016 Tri-lingual Entity Discovery and Linking (EDL) tasks.
Our proposed method has yielded strong performance in all of these examined tasks. 

\section{Related Work}
\label{sec_related_work}

It has been a long history of research involving neural networks (NN). In this section, we briefly review some recent NN-related research work in NLP, which may be relevant to our work. 

The success of word embedding \cite{mikolov2013distributed} encourages researchers to focus on machine-learned representation instead of heavy feature engineering in NLP. Using word embedding as the typical feature representation for words, NNs become competitive to traditional approaches in NER. 
Many NLP tasks, such as NER, chunking and part-of-speech (POS) tagging can be formulated as sequence labeling tasks.
In \cite{collobert2011natural}, 
deep convolutional neural networks (CNN) and conditional random fields (CRF) are used to infer NER labels at a sentence level, where they still  use many hand-crafted features to improve performance, such as capitalization features explicitly defined based on first-letter capital, non-initial capital and so on.

Recently, recurrent neural networks (RNNs) have demonstrated the ability in modeling sequences \cite{graves2012neural}. 
\newcite{huang2015bidirectional} built on the previous CNN-CRF approach by replacing CNNs with bidirectional Long Short-Term Memory (B-LSTM). 
Though they have reported improved performance, they employ heavy feature engineering in that work, most of which is language-specific. There is a similar attempt in \cite{rondeau2016lstm}, where a full-rank CRF is used. 
CNNs are used to extract character-level features automatically in \cite{dos2015boosting}. 

Gazetteer is a list of names grouped by the pre-defined categories an NER system is targeting at. 
Gazetteer is shown to be one of the most effective external knowledge sources to improve NER performance \cite{tjong2003introduction}. Thus,  gazetteer is widely used in many NER systems.  
In \cite{chiu2016named}, state-of-the-art performance on a popular NER task, i.e., CoNLL2003,  is achieved by incorporating  a large gazetteer. 
Different from previous ways to use a set of bits to indicate whether a word is in gazetteer or not, 
they have encoded a match in BIOES (Begin, Inside, Outside, End, Single) annotation, 
which captures positional information. Their models also make advantage of word embeddings, character-level CNNs and CRFs. 

Interestingly enough, none of these recent successes in NER was achieved by a vanilla RNN. Rather, 
these successes are often established by some the sophisticated models combining CNNs, LSTMs and CRFs in certain ways. 
In this paper, based on recent work in  \cite{zhang2015fixed} and \cite{zhang2016compact},
we propose a novel but simple solution to NER by applying DNN on top of FOFE-based features.
This simpler approach can achieve performance very close to state-of-the-art on various NER and mention detection tasks, without using any external knowledge or feature engineering.

\begin{figure*}[t]
\centering
\includegraphics[width=0.8\linewidth]{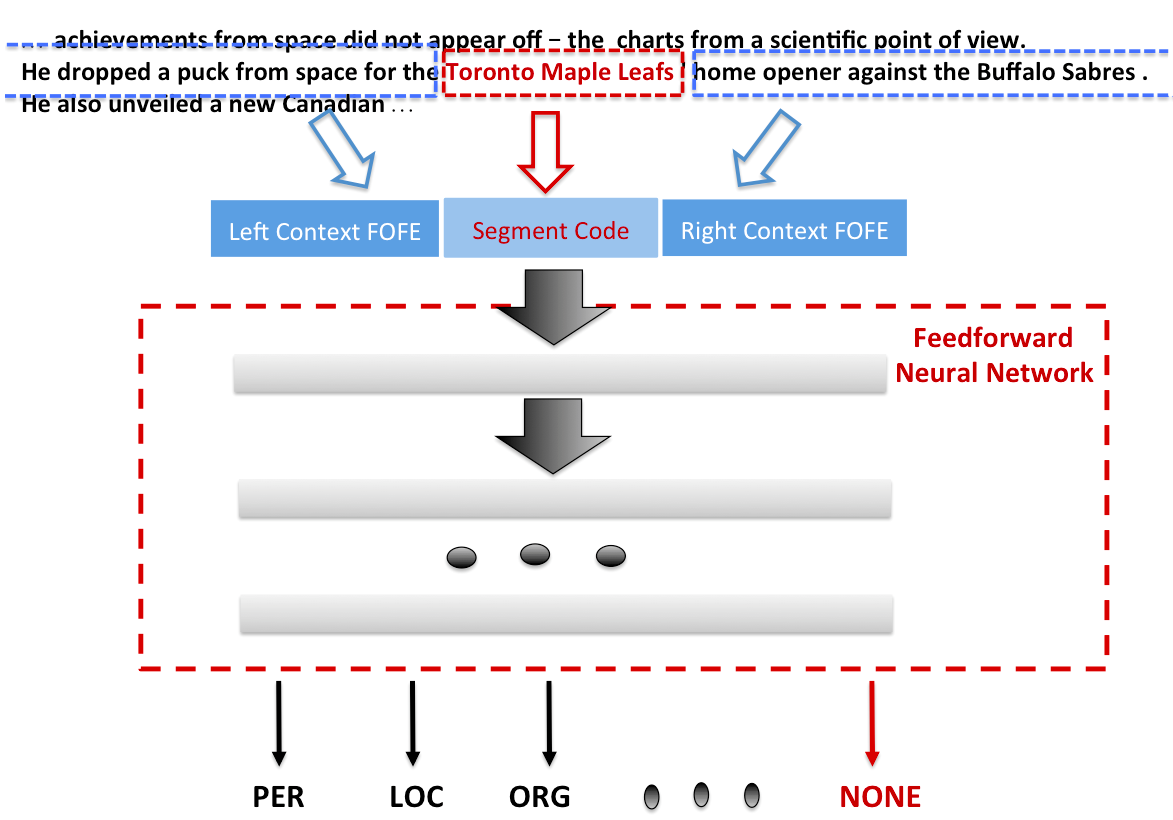}
\caption{Illustration of the local detection approach for NER using FOFE codes as input and a feedforward neural network as model. The window currently examines the fragment of {\it Toronto Maple Leafs}. The window will scan and scrutinize all fragments up to $K$ words. }
\label{Fig:FOFE-NER-diagram}
\end{figure*}

\section{Preliminary}

In this section, we will briefly review some background techniques, which are important to our proposed NER and mention detection approach. 

\subsection{Deep Feedforward Neural Networks}

It is well known that feedforward neural network is a universal approximator under certain conditions \cite{hornik1991approximation}.
A feedforward neural network is a weighted graph with a layered architecture.
Each layer is composed of several nodes, including a bias node whose value is always 1.
Successive layers are fully connected. The nodes in each layer
take as input the values of the nodes in the previous layer, and
compute a function of those values through the connection weights as its
output.

Formally, let $x_{n,j}$ denote the value of the $j$-th node in the
$n$-th layer and $W^n_{i,j}$ denote the weight of the connection from
$x_{n,i}$ to $x_{n+1,j}$. Then
\begin{equation}
z_{n+1,j} = \sum_i W^n_{i,j} x_{n,i}
\end{equation}
\begin{equation}
x_{n+1,j} = \sigma\left(z_{n+1,j}\right)
\end{equation}
where $\sigma$ is the activation function, generally chosen to be sigmoid:
\begin{equation}
\sigma(x) = \frac{1}{1 + e^{-x}}
\end{equation}
or rectified linear unit (ReLU):
\begin{equation}
\sigma(x) = \max(0, x).
\end{equation}
For classification tasks, the outputs are normalized into a probability distribution by the so-called {\it soft-max} function, where the $i$-th node is computed as follows:
\begin{equation}
\sigma(x_i) =  \frac{\exp(x_i)}{\sum_{j} \exp(x_j)}.
\end{equation}

An NN can learn by adjusting its weights in a process called back-propagation. 
Suppose that we have already calculated the outputs given by an NN for any input.
Let $E(y,t)$ be an error metric that measures how incorrect the output
$y$ is with respect to the expected target output $t$. For each weight
in each layer, we may calculate:
\begin{equation}
\frac{\partial E}{\partial W^n_{i,j}} = \frac{\partial E}{\partial
	\sigma} \frac{\partial \sigma}{\partial z_{i+1,j}} \frac{\partial
	z_{i+1,j}}{\partial W^n_{i,j}}.
\end{equation}
Each weight may be adjusted to slowly reduce this error for each training
example, and hence the NN learns to fit the input and the output. 
This is accomplished by the following the update rule, where $\alpha$ is called the learning rate:
\begin{equation}
W^n_{i,j} := W^n_{i,j} - \alpha \frac{\partial E}{\partial
	W^n_{i,j}}.
\end{equation}

The learned NN may be used to generalize and extrapolate to new inputs that have not been seen during training.

\subsection{Fixed-size Ordinally Forgetting Encoding}

Feedforward neural network is a fast and powerful computation model. 
However, it requires to use the fixed-size inputs and 
lacks of the ability to capture long-term dependency in sequences. 
Because most NLP problems involves variable-length sequences of words, 
RNNs/LSTMs are more popular than regular feedforward NNs in dealing with these problems. 
The simple encoding method, called Fixed-size Ordinally Forgetting Encoding (FOFE), 
originally proposed in \cite{zhang2015fixed}, nicely overcomes the limitations of DNNs because it 
can uniquely encode a variable-length sequence of words into a fixed-size representation without lossing information. 

Give a vocabulary $V$ consisting of $|V|$ distinct words, each word can be represented by a one-hot vector. 
FOFE mimics bag-of-words (BOW) but incorporates a forgetting factor to capture positional information.
It encodes any sequence of variable length composed by words in $V$. 
Let $S = {w_1, w_2, w_3, ... , w_T}$ denote a sequence of $T$ words from $V$, 
and $\bm{e_t}$ be the one-hot vector of the $t$-th word in $S$, where $1 \leq t \leq T$.
The FOFE of each partial sequence $\bm{z_t}$ from the first word to the $t$-th word is recursively defined as:
\begin{equation}
\bm{z_t}=
\begin{cases}
\bm{0}, & \text{if}\ t = 0 \\
\alpha \cdot \bm{z_{t - 1}} + \bm{e_t}, & \text{otherwise}
\end{cases}  \label{eq_FOFE_formula}
\end{equation}
where the constant $\alpha$ is called forgetting factor, and it is chosen picked between $0$ and $1$ exclusively. 
Obviously, the size of $\bm{z_t}$ is $|V|$, and it is irrelevant to the length of original sequence, $T$.

Let us use a simple example to illustrate how to use FOFE to decode a sequence. Assume that we 
have three words in our vocabulary, e.g. A, B, C, 
whose one-hot representations are $[1, 0, 0]$, $[0, 1, 0]$ and $[0, 0, 1]$ respectively. 
When calculating from left to right, the FOFE for the sequence "ABC" is $[{\alpha}^2, {\alpha}, 1]$， 
and that of "ABCBC" is $[{\alpha}^4, {\alpha} + {\alpha}^3, 1 + {\alpha}^2]$.

According to \cite{zhang2015fixed}, the word sequences can be unequivocally recovered from their FOFE representations.
The uniqueness of FOFE representation is theoretically guaranteed by the following two theorems: 
\begin{theorem}
	If the forgetting factor $\alpha$ satisfies $0 < \alpha \leq 0.5$, FOFE is unique for any countable vocabulary $V$ and any finite value $T$.
\end{theorem}
\begin{theorem}
	For $0.5 < \alpha < 1 $, given any finite value $T$ and any countable vocabulary $V$,
	FOFE is almost unique everywhere, except only a finite set of countable choices of $\alpha$.
\end{theorem}
Though in theory uniqueness is not guaranteed when $\alpha$ is chosen from $0.5$ to $1$, 
in practice the chance of hitting such scenarios is extremely slim, realistically almost impossible due to quantization errors in the system. 
Furthermore, in natural languages, normally a word does not appear repeatedly within a near context. 
Simply put, FOFE is capable of uniquely encoding any sequence of arbitrary length, serving as a fixed-size but theoretically lossless representation for any sequence.

\subsection {Character-level Models in NLP}
\label{subsec_char_feature}

Recently, as shown in \cite{kim2015character}, it may be beneficial to model morphology in the character level since this may provide some additional advantages in dealing with unknown or out-of-vocabulary (OOVs) words in a language.

The above FOFE method can be easily extended to model character-level feature in NLP. Any word, phrase or fragment can be viewed as a sequence of characters. In this way, based on a pre-defined set of all possible characters, we may apply the same FOFE method to encode the sequence of characters. This always leads to a fixed-size representation, irrelevant to the number of characters in question. 
For example, a word fragment of ``iFLYTEK'' may be viewed as a 
sequence of seven characters: `i', `F', `L', `Y', `T', `E', `K'.
The FOFE codes of  this type of character sequences are always fixed-sized and they can be directly fed to a feedforward neural network for  morphology  modelling. 

In the literature, convolutional neural networks (CNNs) have been widely used as character-level models in NLP \cite{kim2015character}. 
Let $C$ denote the set of possible characters, and $D$ denote the dimensionality of character embeddings.
A $|C| \times D$ matrix $\bm{M}$ is randomly initialized, where the $i$-th row denotes the vector representation of the $i$-th character in $C$. Given a word or fragment whose spelling is $[c_1, c_2, c_3, ..., c_L]$, 
an $L \times D$ matrix $\bm{C}$ is constructed,  where the $j$-th row is a copy of the row in  $\bm{M}$ corresponding to $c_j$. $\bm{C}$ can be viewed as a single-channel image. 
Let $\bm{F}$ be an $h \times D$ convolution kernel to be learned,  where $h$ denotes the number of used feature maps. 
An intermediate vector $\bm{v}$ of $l-h+1$ elements is generated after $\bm{f}$ sweeps $\bm{m}$. Each component in $\bm{v}$,  $v_k$,  is computed as:
\begin{equation}
	\label{eq:conv}
	v_k = \sigma(Trace(\bm{F}\bm{C}[k:k+h]))
\end{equation}
where $\sigma$ is either sigmoid or ReLU.
The output  $y$ of this kernel is given by:
\begin{equation}
	\label{eq:max-pool}
	y = \max(v_1, v_2, ..., v_{l-h+1})
\end{equation}

If there are $N$ groups of kernels, each of which has $n_1$, $n_2$, $n_3$, ... , ${n_{|N|}}$ kernels respectively,
following Eqs. (\ref{eq:conv}) and (\ref{eq:max-pool}), the final representation from the character CNN for this word or fragment is a vector of length $\sum_{i=1}^{|N|} n_{i}$.
	
\section{FOFE-based Local Detection for NER}

As described above, our FOFE-based local detection approach for NER, called  \textbf{FOFE-NER} hereafter, 
is motivated by the way how human actually infers whether a word segment in text is an entity or mention, where  
the entity types of the other entities in the same sentence is not a must.
Particularly, the dependency between adjacent entities is fairly weak in NER problems. 
Whether a fragment is an entity or not, and what class it may belong to, largely depend on
the internal structure of the fragment itself as well as the left and right contexts  in which it appears. 
To a large extent, the meaning and spelling of the underlying fragment 
are informative to distinguish named entities from the rest of the text. 
Contexts play a very important role in NER or mention detection 
when it involves multi-sense words/phrases or out-of-vocabulary (OOV) words. 

As shown in Figure \ref{Fig:FOFE-NER-diagram}, our proposed \textbf{FOFE-NER} method will examine all possible fragments in text (up to a certain length) one by one. For each fragment, it uses the FOFE method to fully encode the underlying fragment itself, its left context and right context into some fixed-size representations, which are in turn fed to a multi-layer feedforward neural network to predict whether the current fragment is not a valid entity mention ({\it NONE}), or its correct entity type ({\it PER}, {\it LOC}, {\it ORG} and so on). This method is appealing because the FOFE codes serves as a theoretically lossless representation of the hypothesis and its full contexts and the multi-layer neural networks are used as a universal approximator to map from text to the entity labels. 


In this work, we use FOFE to explore both word-level and character-level features for each fragment and its contexts. 

\subsection{Word-level Features}

\textbf{FOFE-NER} generates several word-level features for each fragment hypothesis and its left and right contexts as follows:

\begin{itemize}
	\item Bag-of-word vector of the fragment.  For the example in Figure \ref{Fig:FOFE-NER-diagram}, it is a bag-of-word vector of 'Toronto', 'Maple' and 'Leafs'. 
	\item FOFE code for left context including the fragment. In Figure \ref{Fig:FOFE-NER-diagram}, it is the FOFE code of the word sequence of ``{\it ... puck from space for the Toronto Maple Leafs  }".
	\item FOFE code for left context excluding the fragment. In Figure \ref{Fig:FOFE-NER-diagram}, it is the FOFE code of the word sequence of ``{\it ... puck from space for the}".	
	\item FOFE code for right context including the fragment. In Figure \ref{Fig:FOFE-NER-diagram}, it is the FOFE code of the word sequence of ``{\it   ... against opener home ' Leafs  Maple  Toronto}".
	\item FOFE code for right context excluding the fragment. In Figure \ref{Fig:FOFE-NER-diagram}, it is the FOFE code of the word sequence of ``{\it ... against opener home ' }".
\end{itemize}

Moreover, all of the above word features are computed for both case-sensitive words in raw text as well as case-insensitive words in normalized lower-case text. These FOFE codes are projected to lower-dimension dense vectors based on two projection matrices, ${\bf W}_s$ and ${\bf W}_i$, for case-sensitive and case-insensitive FOFE codes respectively. These two projection matrices are initialized by word embeddings trained by {\it word2vec}, and fine-tuned during the learning of the neural networks. 

Due to the recursive computation of FOFE codes in eq.(\ref{eq_FOFE_formula}), all of the above FOFE codes can be jointly computed for one sentence or document in a very efficient manner. 


\subsection{Character-level Features}

On top of the above word-level features, we also augment character-level features for the underlying segment hypothesis to further model its morphological structure. For the example in Figure \ref{Fig:FOFE-NER-diagram}, the current fragment, {\it Toronto Maple Leafs}, is considered as a sequence of case-sensitive characters, i.e. ``\{ 'T', 'o', ..., 'f' , 's'  \}'', we then add the following character-level features for this fragment:

\begin{itemize}
	\item Left-to-right FOFE code of the character sequence of the underlying fragment. That is the FOFE code of the sequence, ``{\it  'T', 'o', ..., 'f' , 's' }''.
	
\item Right-to-left FOFE code of the character sequence of the underlying fragment. That is the FOFE code of the sequence, ``{\it 's' ,  'f' , ..., 'o',  'T' }''.
\end{itemize}

These case-sensitive character FOFE codes are also projected by another character embedding matrix, which is randomly initialized and fine-tuned during model training. 

Alternatively, we may use the character CNNs, as described in Section \ref{subsec_char_feature}, to generate character-level features for each fragment hypothesis as well.

\section{Training and Decoding Algorithm}

Obviously, the above \textbf{FOFE-NER} model will take each sentence of words, $S = [w_1, w_2, w_3, ..., w_m]$, as input, and examine all continuous sub-sequences $[w_i, w_{i+1}, w_{i+2}, ..., w_{j}]$ up to $n$ words in $S$ for  possible entity types. All sub-sequence longer than $n$ words are considered as non-entity in this work. 

When we train the model, based on the entity labels of all sentences in the training set, we will generate many sentence fragments up to $n$ words. These fragments fall into three categories:
\begin{itemize}
	\item Exact-match with an entity label, e.g., the fragment  ``{\it Toronto Maple Leafs}'' in the previous example. 
	\item Partial-overlap with an entity label, e.g., ``{\it for the Toronto}''.
	\item Disjoint with all entity label, e.g. ``{\it from space for}''.
\end{itemize}

For all exact-matched fragments, we generate the corresponding outputs based on the types of the matched entities in the training set. For both partial-overlap and disjoint fragments, we introduce a new output label, {\bf NONE}, to indicate that these fragments are not a valid entity. Therefore, the output nodes in the neural networks contains all entity types plus a rejection option denoted as {\bf NONE}.

During training, we implement a produce-consumer software design such that 
a thread fetches training examples, compute all FOFE codes and packs them as a mini-batch while the other thread feeds the mini-batches to neural networks and adjusts the model parameters and all projection matrices. 
Since ``partial-overlap'' and ``disjoint'' significantly outnumber ``exact-match'', they are down-sampled so as to balance the data set. 


During inference, all fragments not longer than $n$ words are all fed to {\bf FOFE-NER} to compute their scores over all entity types. In practice, these fragments can be packed as one mini-batch so that we can compute them in parallel on GPUs. As the NER result, the {\bf FOFE-NER} model will return a subset of fragments only if: i) they are recognized as a valid entity type (not {\bf NONE}); AND  ii) The NN scores exceed a global pruning threshold. 


Occasionally, some partially-overlapped or nested fragments may occur in the above pruned prediction results. We can use one of the following simple post-processing methods to remove overlappings from the final results:

\begin{enumerate}
	\item {\it highest-first}: We check every word in a sentence. If it is contained by more than one fragment in the pruned results, we only keep the one with the maximum NN score and discard the rest. 
   \item {\it longest-first}: We check every word in a sentence. If it is contained by more than one fragment in the pruned results, we only keep the longest fragment and discard the rest. 
\end{enumerate}

Either of these strategies leads to a collection of non-nested, non-overlapping, non-NONE entity labels.

In some tasks, it may  require to label all nested entities. This has imposed a big challenge to the sequence labelling methods. However, the above post-processing can be slightly modified to generate nested entities' labels. In this case, we first run either {\it highest-first} or {\it longest-first} to generate the first round result. For every entity survived in this round, we will recursively run either {\it highest-first} or {\it longest-first} on all entities in the original set, which are completely contained by it. This will generate more prediction results. This process may continue to allow any levels of nesting. For example, for a sentence of ``$w_1$  $w_2$ $w_3$ $w_4$ $w_5$'', 
if the model first generates the prediction results  after the global pruning, as  [``$w_2 w_3$'', PER, 0.7], [``$w_3 w_4$'', LOC, 0.8], [``$w_1 w_2 w_3 w_4$'', ORG, 0.9], if we choose to run {\it highest-first}, it will generate the first entity label as [``$w_1 w_2 w_3 w_4$'', ORG, 0.9]. Secondly, we will run {\it highest-first} on the two fragments that are completely contained by the first one, i.e.,  [``$w_2 w_3$'', PER, 0.7], [``$w_3 w_4$'', LOC, 0.8], then we will generate the second nested entity label as [``$w_3 w_4$'', LOC, 0.8].
Fortunately, in any real NER and mention detection tasks, it is pretty rare to have overlapped predictions in the NN outputs. Therefore, the extra expense to run this recursive post-processing method is minimal.

\section{Experiments}

In this section, we will evaluate the effectiveness of our proposed methods on several popular NER and mention detection tasks, including the CoNLL 2003 NER task and 
TAC-KBP2015 and TAC-KBP2016 Tri-lingual Entity Discovery and Linking (EDL) tasks.
\footnote{We have made our codes available at https://github.com/xmb-cipher/fofe-ner for readers to reproduce the results in this paper.}

\subsection{CoNLL 2003 NER task}

The CoNLL-2003 dataset \cite{tjong2003introduction} consists of newswire from the Reuters RCV1 corpus tagged with four types of named entities: location (LOC), organization (ORG), person (PER), and miscellaneous (MISC).

We have investigated the performance of our method on the CoNLL-2003 dataset by using different combinations of the FOFE features (both word-level and character-level). The detailed comparison results are shown in Table \ref{tbl:feat-cmp:CoNLL03}.  In Table \ref{tbl:nn-cmp:CoNLL03}, we have compared our best performance with some top-performing neural network systems on this task. As we can see from  Table \ref{tbl:nn-cmp:CoNLL03}, our system yields a very strong performance (90.71 in $F_1$ score) in this task, outperforming most of neural network models reported on this dataset. More importantly, we have not used any hand-crafted features in our systems, and all used features (either word or character level) are automatically derived from the data based on the simple FOFE formula. 
In \cite{chiu2016named}, a slightly better performance (91.62 in $F_1$ score) is reported but a customized gazetteer is used in their method.

\begin{table*}[t]
	\centering
	\begin{tabular}{|l|l|l|lll|}
		\hline
		\multicolumn{3}{|c|}{FEATURE} & P & R & F1\\
		\hline\hline
		\multirow{6}{*}{word-level} & 
		     \multirow{3}{*}{case-insensitive} &
			  context FOFE incl. focus word(s) & 86.64 & 77.04 & 81.56 \\
		& &context FOFE excl. focus word(s) & 53.98 & 42.17 & 47.35  \\
		& & BoW of focus word(s) & 82.92 & 71.85 & 76.99  \\ \cline{2-6} 
		   & \multirow{3}{*}{case-sensitive} & 
			  context FOFE incl. focus word(s) & 88.88 & 79.83 &84.12  \\
		& &context FOFE excl. focus word(s) & 50.91 & 42.46 & 46.30  \\
		& & BoW of focus word(s) & 85.41 & 74.95 & 79.84  \\ \hline
		\multirow{2}{*}{char-level} &
		   \multicolumn{2}{l|}{Char FOFE of focus word(s)} & 67.67 & 52.78 & 59.31  \\
		& \multicolumn{2}{l|}{Char CNN of focus word(s)} & 78.93 & 69.49 & 73.91 \\ \hline
		\multicolumn{3}{|l|}{all case-insensitive features} &  90.11 & 82.75 &  86.28  \\ 
		\multicolumn{3}{|l|}{all case-sensitive features} & 90.26 & 86.63 & 88.41 \\ 
		\multicolumn{3}{|l|}{all word-level features} & 92.03 & 86.08 & 88.96  \\ \hline
		\multicolumn{3}{|l|}{all word-level \& Char FOFE features} & 91.68 &  88.54 & \bf 90.08 \\
		\multicolumn{3}{|l|}{all word-level \& Char CNN features} & 91.80 & 88.58 & \bf 90.16 \\ \hline
		\multicolumn{3}{|l|}{all word-level \& all char-level features}  & 93.29 &  88.27 &  \bf 90.71  \\
		\hline
	\end{tabular}
	\caption{Effect of various FOFE feature combinations on the CoNLL2003 test data.}
	\label{tbl:feat-cmp:CoNLL03}
\end{table*}

\begin{table*}[h]
	\centering
	\begin{tabular}{|l|lllll|l|}
		\hline
		     & word & char & gaz & cap & pos & F1 \\
		\hline\hline
		\cite{collobert2011natural} & \cmark & \xmark & \cmark & \cmark & \xmark & 89.59  \\
		\cite{huang2015bidirectional} &\cmark & \cmark & \cmark & \cmark & \cmark & 90.10 \\
		\cite{rondeau2016lstm}  & \cmark & \xmark & \cmark & \cmark & \cmark & 89.28 \\
		\cite{chiu2016named} & \cmark & \cmark & \cmark & \xmark & \xmark & {\bf 91.62} \\
		\hline \hline
		{\bf this work} & \cmark & \cmark & \xmark & \xmark & \xmark & {\bf 90.71} \\
		\hline
	\end{tabular}
	\caption{Performance ($F_1$ score) comparison among various neural models reported on the CoNLL dataset, and the different features used in these methods.}
	\label{tbl:nn-cmp:CoNLL03}
\end{table*}

\subsection{KBP2015 EDL Task}

Given a document collection in three languages (English, Chinese and Spanish) as input, the KBP2015  tri-lingual EDL task \cite{kbpoverview2015}  requires to automatically identify entity mentions from a
source collection of textual documents in multiple
languages (English, Chinese and Spanish), and 
classify them into one of the following pre-defined
five types: Person (PER), Geo-political Entity
(GPE), Organization (ORG), Location (LOC)
and Facility (FAC), and link them to an existing
English Knowledge Base (KB), and cluster
mentions for those NIL entities that do not have
corresponding KB entries.

\begin{table}
	\resizebox{\linewidth}{!}{
	\centering
	\begin{tabular}{|l|lll|lll|}
		\hline
		\ & \multicolumn{3}{c|}{2015 track best} & \multicolumn{3}{c|}{ours} \\
		\cline{2-7}
		\ & $P$ & $R$ & $F_{1}$ & $P$ & $R$ & $F_{1}$ \\
		\hline \hline
		Trilingual & 75.9 & 69.3 & 72.4 & 78.3 & 69.9 & \bf 73.9 \\
		English & 79.2 & 66.7 & \bf 72.4 & 77.1 & 67.8 & 72.2 \\
		Chinese & 79.2 & 74.8 & \bf 76.9 & 79.3 & 71.7 & 75.3 \\
		Spanish & 78.4 & 72.2 & 75.2 & 79.9 & 71.8 & \bf 75.6 \\
		\hline
	\end{tabular}}
	\caption{Entity Discovery Performance of our method on the KBP2015 EDL evaluation data, with comparison to the best system in KBP2015 official evaluation.}
	\label{Table_KBP2015_ED}
\end{table}

As shown in Table \ref{Table_KBP2015_ED}, our FOFE-based local detection method has obtained pretty strong performance in the KBP2015 dataset. The overall trilingual entity discovery performance is slightly better than the best system participated in the official KBP2015 evaluation, with $73.9$ vs. $72.4$ as measured by $F_1$ scores.

\subsection{KBP2016 EDL task}

In KBP2016, the trilingual EDL task is extended to to detect nominal mentions of all 5 entity types for all three languages. In our experiments, for simplicity, we just treat nominal mention types as some extra entity types and detect them along with named entities together with a single model.  
We have evaluated our proposed FOFE-based local detection method for Entity Discovery in KBP2015 dataset and we have used this method to participate the KBP2016 official tri-lingual EDL evaluation. In the following, we will report the performance of our method on these KBP EDL tasks.

\subsection{Training Data}

For the KBP2015  trilingual EDL task, we make use of the following data sets as our training data to learn the NER and mention detection models. 

\begin{itemize}
	\item \textbf{Training and evaluation data in KBP2015}: In previous year's competition, 335 English documents, 313 Chinese documents and 296 Spanish documents were annotated for training and evaluation, totalling 944 documents. In this data set, all five named mention types (PER, ORG, GPE, LOC, FAC) and only one nominal mention type (PER) are labelled. In KBP2016, nominal mention has been expanded to all 5 classes of named entities.
	
	\item \textbf{Machine-labeled Wikipedia}: When terms or names are first mentioned in a Wikipedia article they are often linked to the corresponding Wikipedia page by hyperlinks, which clearly highlights the possible named entities with well-defined boundary in the text. We have developed a program to automatically map these hyperlinks into KBP annotations by exploring the infobox (if existing) of the destination page and/or examining the corresponding Freebase types. Nominal mentions are not labelled by this approach. In this way, we have created a fairly large amount of  weakly-supervised trilingual training data for the KBP2016 EDL task.
	
	\item \textbf{iFLYTEK's in-house dataset}: The iFLYTEK Research has generously shared with us about 10,000 in-house English and Chinese labeled documents \cite{kbp2016iflytek}.  These documents are internally labelled by iFLYTEK using some annotation rules similar to the KBP 2016 guidelines.
	
\end{itemize}

Additionally, when we generate the machine-labeled data from Wikipedia, we have also created a large gazetteer using the titles of Wikipedia pages and Freebase nodes. 
We have used the gazetteer-related features for the KBP2016 EDL task.

\subsection{Data Preprocessing}

Data from both KBP2015 and KBP2016 are in the  XML format. 
Our preprocessing tools only extract text surrounded by two adjacent XML tags for later stages 
since XML tags tend to be metadata and irrelevant to our task. 
The values of all author attributes are extracted from all post tags, which are directly labeled as {\it PER}.
The extracted text is sent to the Stanford CoreNLP toolkit for sentence splitting and tokenization.  
All words containing digits are mapped to several pre-defined tokens, e.g. $\langle number \rangle$, $\langle date \rangle$, using some regular expression matches.   

\subsection{Hyperparameter optimization}

We perform grid search on several hyper-parameters, 
including initial learning, mini-batch size, initial dropout, 
number of layers, size of hidden layer, number of epochs, on the held-out validation set.
Each hyper-parameter typically has 3 to 5 options during the grid search. 

Here we summarize the set of hyper-parameters used in our experiments:
\begin{itemize}
\item \textbf{KBP series}:
	i) {\it Number of epochs}: we normally run 256 epochs if the iFLYTEK data is not used in training. Otherwise, we only run 64 epochs. ii) {\it Learning rate}: it is initially set to 0.128 and it is gradually decreased by multiplying a number at the end of every epoch so that  it reaches 1/16 of the initial value at the end of the whole training process; iii) {\it Dropout rate}: it is initially set to 0.4 and it is slowly decreased in the training until it reaches 0.1 at the end. iv) {\it Network structure}: we use a feedforward fully-connected structure of 3 hidden layers, each of which has 512 hidden nodes. The ReLU activation function is used. The network weights are randomly initialized based on a uniform distribution between $-\sqrt{\frac{6}{N_i + N_o}}$  and $\sqrt{\frac{6}{N_i + N_o}}$ \cite{glorot2011deep}. v) {\it Embedding matrices}: 
	case-sensitive and case-insensitive word embeddings of 128 dimensions for three languages are pre-trained from English Gigaword, Chinese Wikipeida and Spanish Gigaword using the {\em word2vec} tool \cite{mikolov2013distributed}. Character embeddings have 64 dimensionare and they are randomly initialized. vi) We normally split the available training data into training, validation and evaluation sets in a ratio of 90:5:5.
\item \textbf{CoNLL2003}:
	Similar to \textbf{KBP series} except that i) Word embeddings are of 256 dimensions and trained from Reuters RC1. ii) 128 epochs are run instead.
	iii) We stick to the official data train-dev-test partition.  
\end{itemize}

\subsection{Effect of various training data}

In our first set of experiments, we investigate the effect of using different training data sets on the final entity discovery performance. 
Different training runs are conducted  on different combinations of the aforementioned data sources.
In Table \ref{tbl:kbp2016-EDL1}, we have summarized the official English entity discovery results from three systems we submitted to KBP2016 EDL1 evaluation. The first system, using only the KBP2015 data to train the model, has achieved 0.693 in $F_1$ score in the official KBP2016 English evaluation data. After adding the weakly labelled data, WIKI, we can see the entity discovery performance is improved to 0.707 in  $F_1$ score. Finally, we can see that it yields the best performance by using the KBP2015 data and the iFLYTEK in-house data sets to train our models, giving 0.731 in $F_1$ score.

\begin{table}[t]
	\centering
	\begin{tabular}{|c|ll|l|}
		\hline
		training data  &  P   &  R  &  $F_1$ \\ \hline \hline
			KBP2015 &  0.818 & 0.600 & 0.693 \\
			KBP2015 + WIKI  &   0.859 & 0.601 & 0.707 \\	
			KBP2015 + iFLYTEK &  0.830 & 0.652 & \bf 0.731 \\
      \hline
	\end{tabular}
	\caption{Entity discovery performance (English only) in KBP2016 EDL1 evaluation window is shown as a comparison of three models trained by different combinations of training data sets. }
	\label{tbl:kbp2016-EDL1}	
\end{table}

\subsection{The official performance in KBP2016 EDL evaluation}

After fixing some system bugs, we have used both the KBP2015 data and iFLYTEK data to re-train our models for three languages and finally submitted three systems to the final KBP2016 EDL2 evaluation. The official results of two systems are summarized in Table \ref{tbl:kbp2016}. In our systems, we treat all nominal mentions as special types of named entities and both named and nominal entities are recognized using one model. Here we have broken down the system performance according to different languages and categories of entities (named or nominal). In RUN1, we have submitted our best NER system, achieving about 0.718 in $F_1$ score in the KBP2016 trilingual EDL track. This  is a very strong performance among all KBP2016 participating teams. In RUN3, we have submitted system fusion results by combining our results with the best results from another KBP2016 participating team using CNNs and RNNs \cite{kbp2016iflytek}. The overall trilingual $F_1$ score is improved to 0.754.  It is worth to note that we have obtained a pretty high recall rate, about 0.735, after the system combination because the NER methods used by these two systems are quite complementary.  

\begin{table*}[t]
	\centering
	\begin{tabular}{|l|lll|lll|lll|}
		\hline
		 \multirow{2}{*}{LANG}  &
			\multicolumn{3}{c|}{NAME} & \multicolumn{3}{c|}{NOMINAL} & \multicolumn{3}{c|}{OVERALL} \\
		 \ & P & R & F1 & P & R & F1 & P & R & F1 \\
		\hline \hline
		& \multicolumn{9}{c|}{RUN1 (our official ED result in KBP2016 EDL2)} \\ 
		\hline
		  	ENG & 0.898 & 0.789 & 0.840 & 0.554 & 0.336 & 0.418 & 0.836 & 0.680 & 0.750 \\
		   CMN & 0.848 & 0.702 & 0.768 & 0.414 & 0.258 & 0.318 & 0.789 & 0.625 & 0.698 \\
		   SPA & 0.835 & 0.778 & 0.806 & 0.000 & 0.000 & 0.000 & 0.835 & 0.602 & 0.700 \\
		   ALL & 0.893 & 0.759 & 0.821 & 0.541 & 0.315 & 0.398 & 0.819 & 0.639 & {\bf 0.718} \\
		\hline
		& \multicolumn{9}{c|}{RUN3 (system fusion of RUN1 with the best system in \cite{kbp2016iflytek})} \\
		\hline
		 ENG & 0.857	& 0.876 & 0.866 & 0.551 & 0.373 & 0.444 & 0.804 & 0.755 & 0.779 \\
		 CMN & 0.790 & 0.839 & 0.814 & 0.425 & 0.380 & 0.401 & 0.735 & 0.760 & 0.747 \\
		 SPA & 0.790 & 0.877 & 0.831 & 0.000 & 0.000 & 0.000 & 0.790 & 0.678 & 0.730 \\
		 ALL & 0.893 & 0.759 & 0.821 & 0.541 & 0.315 & 0.398 & 0.774 & {\bf 0.735} & {\bf 0.754} \\
		\hline 
	\end{tabular}
	\caption{Official entity discovery performance of our methods on KBP2016 trilingual EDL track.}
	\label{tbl:kbp2016}
\end{table*}

\section{Conclusion}

In this paper, we have proposed a new local detection based approach, which rely on the recent fixed-size ordinally forgetting encoding (FOFE) method to fully encode each fragment and its left/right contexts into a fixed-size representation. Afterwards, a simple feedforward neural network is used to reject or predict entity label for each individual fragment. The proposed method has been evaluated in several popular NER and mention detection tasks, including the CoNLL 2003 NER task and  TAC-KBP2015 and TAC-KBP2016 Tri-lingual Entity Discovery and Linking (EDL) tasks. Our methods have yielded pretty strong performance in all of these examined tasks. 

Obviously, this FOFE-based local detection approach can be easily extended to tackle many other NLP tasks, such as chunking, POS tagging, entity linking, semantic parsing. We will report our progresses in these new tasks in the future.  

\bibliography{fofe4ner}
\bibliographystyle{acl2016}

\end{document}